\documentclass[letterpaper, 10 pt, conference]{ieeeconf}  

\usepackage{amsmath}
\usepackage{amssymb}
\usepackage{graphicx}
\usepackage{booktabs}
\usepackage{multirow}
\usepackage{array}
\usepackage{tabularx}
\usepackage[caption=false,font=footnotesize]{subfig}
\usepackage{cite}

\usepackage[svgnames]{xcolor}
\usepackage{soul} 
\usepackage[colorlinks=true, linkcolor=blue, citecolor=red, urlcolor=magenta]{hyperref}  

\usepackage{tikz}
\newcommand\copyrightnotice[1]{
    \begin{tikzpicture}[remember picture,overlay]
    \node[anchor=north,xshift=125,yshift=-12pt] at (current page.north) {\parbox{\dimexpr0.45\textwidth-\fboxsep-\fboxrule\relax}{\scriptsize #1}};
    \end{tikzpicture}
}

\IEEEoverridecommandlockouts                              

\title{\LARGE \bf
Potential-Field Action Representation for Reinforcement Learning in Contact-Rich Manipulation

}

\author{Xinyu Liu, Gökhan Solak and Arash Ajoudani
\thanks{This paper was supported by the European Union Horizon Projects TORNADO (Grant GA 101189557)  and by the Italian Ministry of University and Research (MUR) under the Fondo Italiano per la Scienza (FIS), call FIS 3, project EPIC with code FIS-2024-02654.}
\thanks{Xinyu Liu is with Human-Robot Interfaces and Interaction Lab, Istituto Italiano di Tecnologia, 16121 Genoa, Italy, and also with the Ph.D. Program of National Interest in Robotics and Intelligent Machines (DRIM), Università di Genova, 16126 Genoa, Italy.
}
\thanks{Gökhan Solak and Arash Ajoudani is with the Human-Robot Interfaces and Interaction Laboratory, Istituto Italiano di Tecnologia, 16163 Genoa, Italy.}
}

\begin{document}

\maketitle
\thispagestyle{empty}
\pagestyle{empty}

\begin{abstract}

Model-free reinforcement learning can acquire contact-rich robotic manipulation skills through trial-error interaction, but it often requires the policy to learn both task strategy and low-level motion generation from data. 
In this setting, the action representation is critical because it determines how policy outputs are converted into robot motion and therefore influences both exploration and physical execution. Direct Cartesian motion-command interfaces are widely used, but they require the policy to generate motion commands at every decision step, coupling task-level adaptation with continuous low-level execution. 
This increases the learning burden, since the policy must also discover commands that are smooth, bounded, and physically suitable for compliant interaction.
We propose reinforcement learning with artificial potential-fields as action representation (PA-RL). Instead of commanding motion directly, the policy adapts the parameters of an energy-like potential field which provides a state-dependent guidance direction that is executed through a Cartesian impedance controller. We evaluate the PA-RL framework on peg-in-hole insertion, a representative contact-rich task with nonlinear robot dynamics and discontinuous contact transitions. Simulation experiments compare PA-RL with direct Cartesian velocity, Cartesian pose, and variable-impedance action spaces trained with the same RL algorithm. PA-RL is the first and only method train a policy to reach a \(100\%\) evaluation success rate within given time while the best baseline reaches \(92.6\%\). Compared with the best-performing baseline, PA-RL reduced the joint-torque variation by \(55.4\%\) and Cartesian acceleration variation by \(70.8\%\), respectively, without explicit motion quality penalties in the reward. The simulation-trained policy further completed all \(9/9\) real-robot insertions without policy fine-tuning, demonstrating deployment feasibility of the learned potential-field interface.

\end{abstract}

\copyrightnotice{\copyright 2026 IEEE. 
This work has been submitted to the IEEE for possible publication.
Copyright may be transferred without notice, after which this version may no longer be accessible.}

\section{INTRODUCTION}
\label{introduction}

\begin{figure}[bt]
    \centering
    
        
    \includegraphics[width=0.9\linewidth]{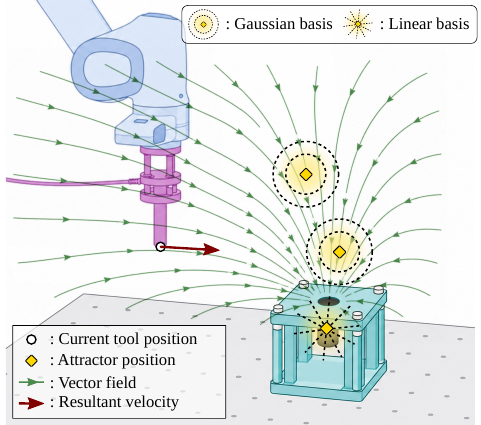}
    
    \caption{Overview of the PA-RL framework for the peg-in-hole task. The policy action consists of the positions and parameters of several attractors with linear and gaussian bases, defining a state-dependent artificial potential field over the goal-relative workspace that is illustrated as a vector field (green) along its gradient. The velocity reference is generated following the negative gradient of the potential field.}
    \label{fig:intro}
\end{figure}

Humanoid robots are expected to operate in human-centered environments and perform diverse manipulation tasks that involve sustained physical interaction, including assembly, maintenance, and tool use.\cite{Gu2026Humanoid, riener2023robots} Such contact-rich manipulation is challenging because robot motion is affected by constrained geometry, nonlinear dynamics, discontinuous contact transitions, and uncertainty in the task state \cite{elguea2023review, zhang2026safe}. Successful execution therefore requires the robot to adapt its high-level strategy to the task and contact conditions while maintaining precise and compliant motion during physical interaction.

Robot control methods, such as operational space control and impedance control, provide effective structures for precise and compliant motion during physical interaction. However, the task-level reference trajectory and interaction strategies are commonly designed manually and require substantial adaptation when task conditions change. 
Reinforcement learning (RL) provides a complementary approach by allowing robots to acquire difficult-to-engineer behaviors through trial-and-error interaction. 
Nevertheless, the learning algorithm and reward function alone do not determine how a learned policy interacts with the robot. The action representation is critical: it defines how policy decisions are translated into executable motion and influences both exploration and physical execution\cite{martin2019variable, bahl2020neural, zhang2024SRL
}. Many RL methods use direct motion-command action representations, including joint commands, pose increments, and velocities. 
These representations are general and flexible, but they couple task-level decision making with low-level motion generation: the same policy output must both select an interaction strategy and produce an executable command. Consequently, variations between consecutive policy decisions can be directly reflected in the robot motion, which may reduce exploration efficiency and execution smoothness during contact.

For contact-rich manipulation, this motivates action representations that encode a spatial motion structure rather than an isolated step-wise motion command. 
Artificial potential fields provide a particularly relevant representation because they encode motion as an energy-like landscape  over the task space. Their negative gradient acts as a virtual force, yielding guidance that is physically interpretable, spatially structured, and smooth over the task space rather than defined only as isolated step-wise commands. However, conventional artificial potential fields\cite{khatib1986potential, Huang2026Globallystableneuralimitation} are commonly designed manually and remain fixed during execution. Their behavior can therefore depend strongly on the selected field parameters, limiting their adaptability.

Therefore, this paper proposes PA-RL that combines reinforcement learning and artificial potential-fields. The core idea is to use a parameterized artificial potential field as the action representation of the RL policy. Based on goal-relative state and force--torque observations, the PA-RL policy adapts the positions, weights, and widths of a set of attractors. These parameters define a state-dependent potential field, as shown in \autoref{fig:intro}, whose negative gradient is used to generate a bounded reference trajectory tracked by a fixed-gain Cartesian impedance controller. Consequently, the policy adapts the spatial structure of the motion, while continuous execution is retained within the potential-field and impedance control structure.

The proposed method is evaluated on peg-in-hole, a representative contact-rich task with nonlinear dynamics and discontinuous contact transitions\cite{liu2025general, Liu2024}. Practically, the precise hole pose is difficult to obtain because of calibration errors, sensing uncertainty, and manufacturing tolerances. Therefore, successful insertion requires the policy to use not only the estimated geometric state but also contact-related observations to correct its motion during execution. These characteristics make peg-in-hole insertion a suitable benchmark for evaluating adaptive policy interfaces.

PA-RL is evaluated against three representative action-space baselines: end-effector velocity (TCP-Vel), incremental end-effector pose (TCP-Pose) and VICES-style variable-impedance that outputs an incremental pose together with stiffness and damping parameters. All methods use the same RL algorithm and reward, isolating the role of the action representation. The results show that the PA-RL improves the learning efficiency, task performance, and motion quality, and that the simulation-trained policy transfers to the real robot without fine-tuning.

The main contributions of this work are:
\begin{itemize}
    \item The PA-RL framework that uses a state-dependent artificial potential field as RL action representation. The framework decouples learned task-level adaptation from continuous reference generation and execution, providing a suitable framework for robot control.
    \item A systematic comparison of PA-RL with TCP-Vel, TCP-Pose, and a VICES-style variable-impedance baseline with the same RL algorithm. 
    The evaluation examines learning efficiency, task performance, and motion quality.
    \item A real-robot deployment of the simulation-trained PA-RL policy on a Franka Emika Panda without policy fine-tuning, demonstrating transfer feasibility and contact-dependent adaptation in peg-in-hole insertion.
\end{itemize}

\section{Related Work}

\subsection{Action Representations for Contact-Rich Manipulation}
RL has been applied to contact-rich manipulation
using a wide range of action representations. 
Common choices include joint torques, joint positions, Cartesian poses, pose increments, and velocities~\cite{elguea2023review, zhang2026safe}. These representations are general and can be used with standard RL algorithms.

Control-inspired action representations incorporate additional structure into this action representation. Variable-impedance approaches allow the policy to regulate task-space motion together with stiffness and damping parameters. For example, VICES studies end-effector variable-impedance control as an action space for constrained and contact-rich tasks~\cite{martin2019variable}. Other methods embed energy-shaping, passivity, or stability-oriented controller structures into learned policies~\cite{khader2020stability,khader2020stable, zhang2024SRL}. These approaches demonstrate that incorporating robot control structure into the policy representation can influence both learning and physical execution.

Structured dynamical-system representations provide another alternative to independent step-wise commands. Dynamic movement primitives (DMPs) represent motion through attractor dynamics and a phase-dependent forcing term, providing a compact representation of temporally evolving trajectories~\cite{ijspeert2013dynamical, davchev2022residual}. Neural Dynamic Policies further embed such differential-equation structures into deep policies for reinforcement and imitation learning, allowing the policy to predict a trajectory level representation rather than raw time-step-wise actions~\cite{bahl2020neural}.

PA-RL emphasizes a different form of structure. Rather than encoding a trajectory through a prescribed phase progression, the policy parameterizes a spatial potential field that is evaluated at the current robot state and updated from geometric and contact-related observations.

\subsection{Artificial Potential Fields for Robot Control}
Artificial potential fields were introduced as a reactive approach to
robot motion generation and obstacle avoidance~\cite{khatib1986potential}. A scalar potential is defined over the task space, and its negative gradient determines a state-dependent motion direction. Attractive and repulsive potential components can be combined to express task objectives and environmental constraints. Their spatial formulation and real-time performance have made artificial potential fields widely applicable to reactive robot motion.

Artificial potential fields can be integrated into robot control systems in different ways. Given a potential function $(U(\mathbf{x}))$, the negative gradient may be interpreted as a virtual force:

\begin{equation}
\mathbf{F}(\mathbf{x})
=
-\nabla_{\mathbf{x}} U(\mathbf{x}).
\end{equation}

It could be converted into a velocity or position reference, or combined with operational-space and impedance controllers.Potential fields have also been incorporated into structured trajectory representations; for example, potential-field terms have been used to provide obstacle avoidance within DMP-generated motion~\cite{park2008movement}. It also embedded within manually structured task procedure like the approach stage in assembly task which allow different motion-generation mechanisms\cite{zhang2025novel}.

The behavior of an artificial potential field depends on the selected potential functions, their spatial locations, and their relative contributions. Classical formulations typically specify these elements manually for a given task and environment. Learning-based methods have therefore investigated stable dynamical systems and potential functions learned from demonstrations~\cite{khansari2011stable}. Such methods can encode demonstrated motion and compliant behavior within a learned spatial structure.

In this study, PA-RL treats the parameters of the artificial potential field as the online action space of an RL policy. The field is reshaped during task execution according to the current state including force--torque observations, rather than learned once as a fixed motion model. The artificial potential field therefore acts as a bridge between adaptive task-level decisions and continuous robot execution.

\section{Methodology}
\label{methodolgy_reinforcement learning}

\begin{figure}[htbp]
    \centering
    \includegraphics[width=1\linewidth]{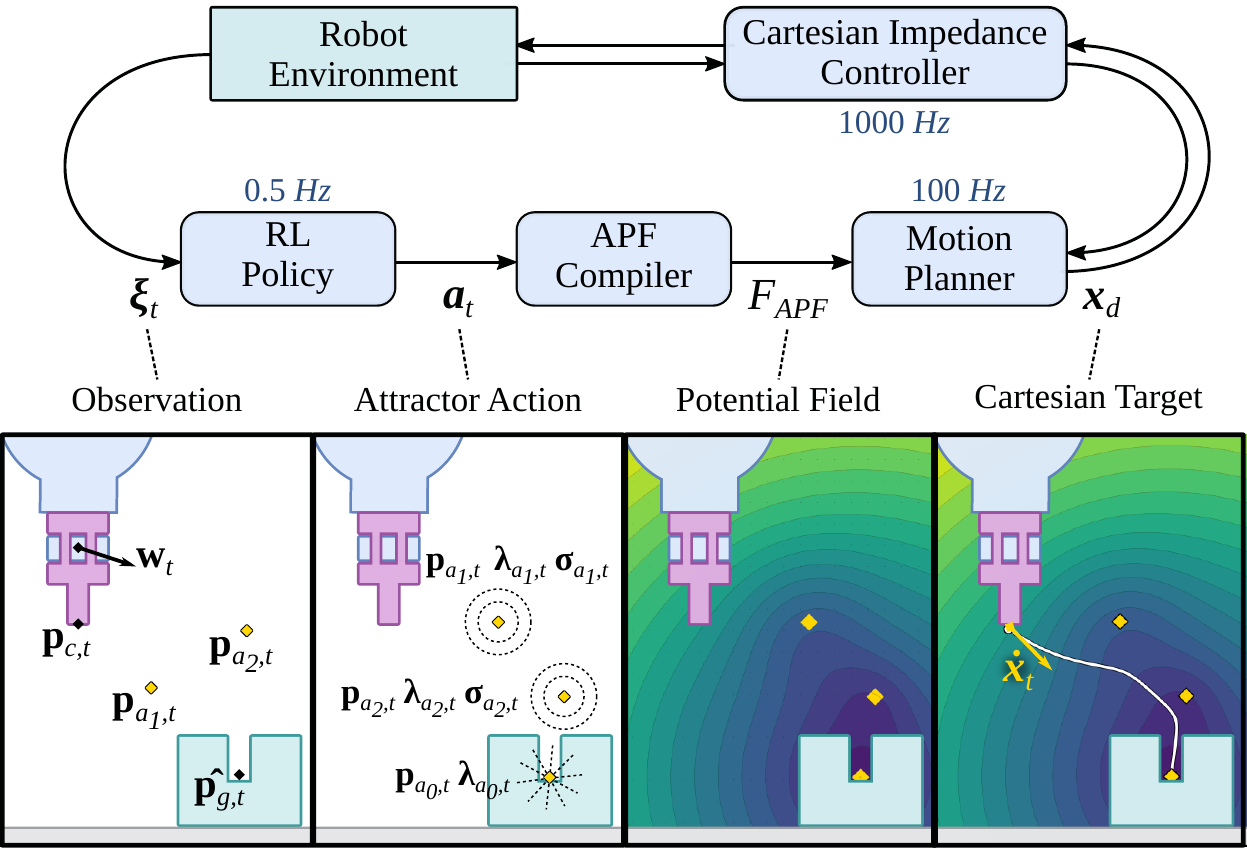}
    \caption{Overview of the PA-RL pipeline. The upper diagram shows the closed-loop execution architecture and the lower panels illustrate a representative peg-in-hole example. Here, \(\mathbf{p}_{c,t}\) and \({\mathbf{p}}_{g,t}\) are peg-tip position and goal position, and force--torque sensor wrench \(\mathbf{w}_t\). The action specifies the attractor offsets \({\mathbf{p}}_{a_{i},t}\), weights \({\mathbf{\lambda}}_{a_{i},t}\), and Gaussian sigma \({\mathbf{\sigma}}_{a_{i},t}\).
    }
    \label{fig:method}
\end{figure}

In the PA-RL framework, as illustrated in \autoref{fig:method}, the policy maps observations to attractor parameters that define a state-dependent artificial potential field. Then, based on the field gradient, a Cartesian reference is generated and tracked by a Cartesian impedance controller. This decouples learned field adaptation from continuous motion generation and robot control.

\subsection{Artifical Potential Field}

An artificial potential field (APF) defines a scalar function over the Cartesian task space whose negative gradient induces a motion direction. Let \(\mathbf{x}\in\mathbb{R}^3\) denote a controlled Cartesian point, and let \(\mathbf{p}_{a_i}\in\mathbb{R}^3\) denote the center of the \(i\)-th basis. We construct the APF as a weighted combination of linear and Gaussian radial bases,
\begin{equation}
U(\mathbf{x})
=
\frac{1}{\sum_i |\lambda_i|}
\sum_i U_i(\mathbf{x}).
\end{equation}
where \(\lambda_i\) is the signed weight of basis \(i\). The normalization prevents the overall field magnitude from scaling directly with the absolute sum of learned weights.

For a linear basis, the potential is defined as
\begin{equation}
U_i^{\mathrm{lin}}(\mathbf{x})
=
\frac{1}{2}\lambda_i
\|\mathbf{x}-\mathbf{p}_{a_i}\|^2 .
\end{equation}
which yields the vector-field contribution
\begin{equation}
\mathbf{f}_i^{\mathrm{lin}}(\mathbf{x})
=
-\nabla_{\mathbf{x}} U_i^{\mathrm{lin}}(\mathbf{x})
=
\lambda_i(\mathbf{p}_{a_i}-\mathbf{x}).
\end{equation}
To increase the expressiveness of the field for contact-rich manipulation, we also include localized nonlinear bases. For a Gaussian radial basis, we use
\begin{equation}
U_i^{\mathrm{gau}}(\mathbf{x})
=
\lambda_i\sigma_i\sqrt{\frac{\pi}{2}}\,
\mathrm{erf}
\left(
\frac{\|\mathbf{x}-\mathbf{p}_{a_i}\|}{\sqrt{2}\sigma_i}
\right).
\end{equation}
which gives
\begin{equation}
\begin{aligned}
\mathbf{f}_i^{\mathrm{gau}}(\mathbf{x})
&=
-\nabla_{\mathbf{x}} U_i^{\mathrm{gau}}(\mathbf{x}) \\
&=
-\lambda_i
\exp\left(
-\frac{\|\mathbf{x}-\mathbf{p}_{a_i}\|^2}{2\sigma_i^2}
\right)
\frac{\mathbf{x}-\mathbf{p}_{a_i}}
{\|\mathbf{x}-\mathbf{p}_{a_i}\|+\epsilon}.
\end{aligned}
\end{equation}
Here, \(\sigma_i\) controls the spatial width of the Gaussian basis and \(\epsilon>0\) avoids numerical singularities near the basis center.

Using \(\mathbf{f}_i(\mathbf{x})\) to denote either the linear \(\mathbf{f}_i^{\mathrm{lin}}(\mathbf{x})\) or the Gaussian \(\mathbf{f}_i^{\mathrm{gau}}(\mathbf{x})\) basis function, the resulting APF vector field is
\begin{equation}
\mathbf{F}_{\mathrm{APF}}(\mathbf{x})
=
\frac{\sum_i \mathbf{f}_i(\mathbf{x})}
{\sum_i |\lambda_i|}.
\end{equation}

In the proposed method, the APF parameters are not manually fixed. Instead, the RL policy outputs the basis parameters as described in the next section. 

\subsection{Reinforcement Learning (RL)}




The RL policy adapts the APF parameters from the current task context. We use one linear attractor near the observed goal and \(N\) Gaussian attractors, resulting in \(N+1\) APF attractors in total. At each environment step \(t\), the policy receives an observation composed of the goal-relative Cartesian state, the measured interaction wrench, and the current attractor offsets:
\begin{equation}
\boldsymbol{\xi}_t
=
\left[
\Delta\mathbf{p}_{g,t}^{\top}
\ \big|\
\mathbf{w}_{t}^{\top}
\ \big|\
\Delta\mathbf{p}_{\mathcal{A},t}^{\top}
\right]^{\top},
\end{equation}
where
\begin{equation}
\Delta\mathbf{p}_{g,t}
=
\widehat{\mathbf{p}}_{g,t}
-
\mathbf{p}_{c,t}
\end{equation}
denotes the displacement from the controlled Cartesian point \(\mathbf{p}_{c,t}\) to the observed task goal \(\widehat{\mathbf{p}}_{g,t}\). The vector \(\mathbf{w}_t\in\mathbb{R}^{6}\) is the measured force--torque signal. The stacked attractor-offset vector is
\begin{equation}
\Delta\mathbf{p}_{\mathcal{A},t}
=
\left[
\Delta\mathbf{p}_{a_0,t}^{\top},
\Delta\mathbf{p}_{a_1,t}^{\top},
\ldots,
\Delta\mathbf{p}_{a_N,t}^{\top}
\right]^{\top},
\end{equation}
where \(a_0\) denotes the linear attractor and \(a_1,\ldots,a_N\) denote the Gaussian attractors. Thus, the observation dimension is $\dim(\boldsymbol{\xi}_t) = 3 + 6 + 3(N+1).$

The policy outputs a normalized action
\begin{equation}
\mathbf{a}_t
=
\left[
\Delta\mathbf{p}_{\mathcal{A},t}^{\top},
\boldsymbol{\lambda}_t^{\top},
\boldsymbol{\sigma}_t^{\top}
\right]^{\top},
\end{equation}
where \(\Delta\mathbf{p}_{\mathcal{A},t}\in\mathbb{R}^{3(N+1)}\) contains the attractor offsets, 
\(\boldsymbol{\lambda}_t\in\mathbb{R}^{N+1}\) contains the signed attractor weights, and
\(\boldsymbol{\sigma}_t\in\mathbb{R}^{N}\) contains the Gaussian widths. The linear attractor does not require a Gaussian width parameter. Therefore, $\mathbf{a}_t \in \mathbb{R}^{5N+4}.$


All policy actions are normalized to \([-1,1]\) and mapped to predefined physical ranges before constructing the APF. The Cartesian position of attractor \(k\) is computed as
\begin{equation}
\mathbf{p}_{a_k,t}
=
\widehat{\mathbf{p}}_{g,t}
+
\Delta\mathbf{p}_{a_k,t},
\qquad
k\in\{0,\ldots,N\}.
\end{equation}
The decoded offsets, signed weights, and Gaussian widths define the APF vector field \(\mathbf{F}_{\mathrm{APF}}(.)\), which is then converted into a bounded Cartesian reference by the trajectory generation module.

The policy is trained using Soft Actor--Critic (SAC)\cite{haarnoja2018sac}, an off-policy maximum-entropy actor--critic algorithm. SAC optimizes a stochastic policy by maximizing the expected discounted return while encouraging exploration through an entropy term:
\begin{equation}
J(\pi)
=
\mathbb{E}_{\pi}
\left[
\sum_{t=0}^{T}
\gamma^t
\left(
r_t
+
\alpha
\mathcal{H}
\left(
\pi(\cdot|\boldsymbol{\xi}_t)
\right)
\right)
\right],
\end{equation}
where \(r_t\) is the task reward, \(\gamma\) is the discount factor, \(\alpha\) is the entropy temperature, and \(\mathcal{H}(\pi(\cdot|\boldsymbol{\xi}_t))\) denotes the policy entropy.

Thus, the learned policy does not directly command torques, forces, or impedance gains. It selects APF parameters, while the generated field is subsequently transformed into a bounded Cartesian reference and executed by the low-level controller.

\subsection{Motion Planning and Control}
\label{Trajectory Planning and Control}

The APF defined above is not applied directly as a Cartesian force. Instead, it is used as a bounded reference generator for the equilibrium point of a Cartesian impedance controller. Given the APF vector field \(\mathbf{F}_{\mathrm{APF}}(.)\), the reference velocity is computed as
\begin{equation}
\dot{\mathbf{x}}_{d,t} =
\mathrm{sat}_{v_{\max}}
\left(
g_v \mathbf{F}_{\mathrm{APF},t}(\mathbf{p}_{c,t})
\right),
\label{eq:bounded_reference_velocity}
\end{equation}
where \(g_v\) is a velocity gain and \(\mathrm{sat}_{v_{\max}}(\cdot)\) limits the velocity norm by \(v_{\max}\). This saturation decouples the magnitude of the commanded Cartesian motion from abrupt changes in the learned APF parameters.

The Cartesian target is then obtained by integrating the bounded reference velocity,
\begin{equation}
\mathbf{x}_{d,t+\Delta t}
=
\mathbf{x}_{d,t} + \Delta t\,\dot{\mathbf{x}}_{d,t}.
\label{eq:target_integration}
\end{equation}
The target is projected into the admissible workspace and constrained to remain within a bounded lead distance from the measured robot position,
\begin{equation}
\mathbf{x}_{d,t}\in\mathcal{X},
\qquad
\|\mathbf{x}_{d,t}-\mathbf{p}_{c,t}\| \le \ell_{\max}.
\label{eq:target_constraints}
\end{equation}
Thus, the APF-policy module produces a bounded moving equilibrium rather than a direct torque, force, or stiffness command.

The generated target is tracked by a Cartesian impedance controller with fixed positive-definite stiffness and damping,
\begin{equation}
\mathbf{F}_{\mathrm{imp},t} =
K_p(\mathbf{x}_{d,t}-\mathbf{p}_{c,t})
+
D_p(\dot{\mathbf{x}}_{d,t}-\dot{\mathbf{p}}_{c,t}).
\label{eq:cartesian_impedance}
\end{equation}
The corresponding joint torques are computed through the manipulator Jacobian,
\begin{equation}
\boldsymbol{\tau}_t
=
J_t^{\top}\mathbf{F}_{\mathrm{imp},t}
+
\boldsymbol{\tau}_{\mathrm{null},t}
+
\boldsymbol{\tau}_{\mathrm{bias},t}.
\end{equation}
where \(\boldsymbol{\tau}_{\mathrm{null},t}\) denotes the null-space posture regulation term and \(\boldsymbol{\tau}_{\mathrm{bias},t}\) accounts for model compensation. Therefore, the learned policy shapes the APF used for trajectory generation, while the robot remains controlled through a bounded reference update and a fixed damped impedance law.

\subsection{Stability and Boundedness}

The learned APF enters the robot controller only through the commanded
Cartesian equilibrium \(\mathbf{x}_{d,t}\). From
\eqref{eq:bounded_reference_velocity}--\eqref{eq:target_constraints}, the
reference velocity and its displacement from the controlled point
\(\mathbf{p}_{c,t}\) are bounded. Thus, abrupt policy-induced APF changes
cannot generate unbounded reference jumps.

Consider the Cartesian impedance storage function
\begin{equation}
V_t =
\frac{1}{2}\dot{\mathbf{p}}_{c,t}^{\top}
M_x(q_t)
\dot{\mathbf{p}}_{c,t}
+
\frac{1}{2}
(\mathbf{p}_{c,t}-\mathbf{x}_{d,t})^{\top}
K_p
(\mathbf{p}_{c,t}-\mathbf{x}_{d,t}).
\end{equation}
For fixed positive-definite \(K_p,D_p\), the damping term in
\eqref{eq:cartesian_impedance} dissipates energy. The energy injected by the
moving reference is bounded by
\begin{equation}
\left|
(\mathbf{p}_{c,t}-\mathbf{x}_{d,t})^{\top}
K_p
\dot{\mathbf{x}}_{d,t}
\right|
\le
\ell_{\max}\|K_p\|v_{\max},
\end{equation}
and the elastic interaction is bounded by
\begin{equation}
\|K_p(\mathbf{x}_{d,t}-\mathbf{p}_{c,t})\|
\le
\|K_p\|\ell_{\max}.
\end{equation}

Therefore, the learned APF does not act as an unbounded force source; it acts as a bounded reference input to a damped impedance system. Under bounded workspace operation and fixed positive-definite impedance gains, the closed-loop system is practically bounded with respect to the commanded Cartesian equilibrium. 
This argument does not imply global asymptotic convergence to the task goal, since local APF minima and obstacle-induced detours may occur. It instead establishes that policy-induced APF changes cannot generate unbounded reference motion or unbounded impedance deflection.

\section{Experiments}
The PA-RL framework is evaluated on the peg-in-hole task with three baselines and they are presented consistently in the order PA-RL, TCP-Vel, TCP-Pose, and VICES. The experiments are organized around learning efficiency, task performance, and motion quality. Finally the PA-RL policy transferred to real-robot to assess deployment feasibility.

\subsection{Task and Training Setup}
\label{sec:task_simulation_setup}

\begin{figure}[htbp]
    \centering
    \includegraphics[width=0.85\linewidth]{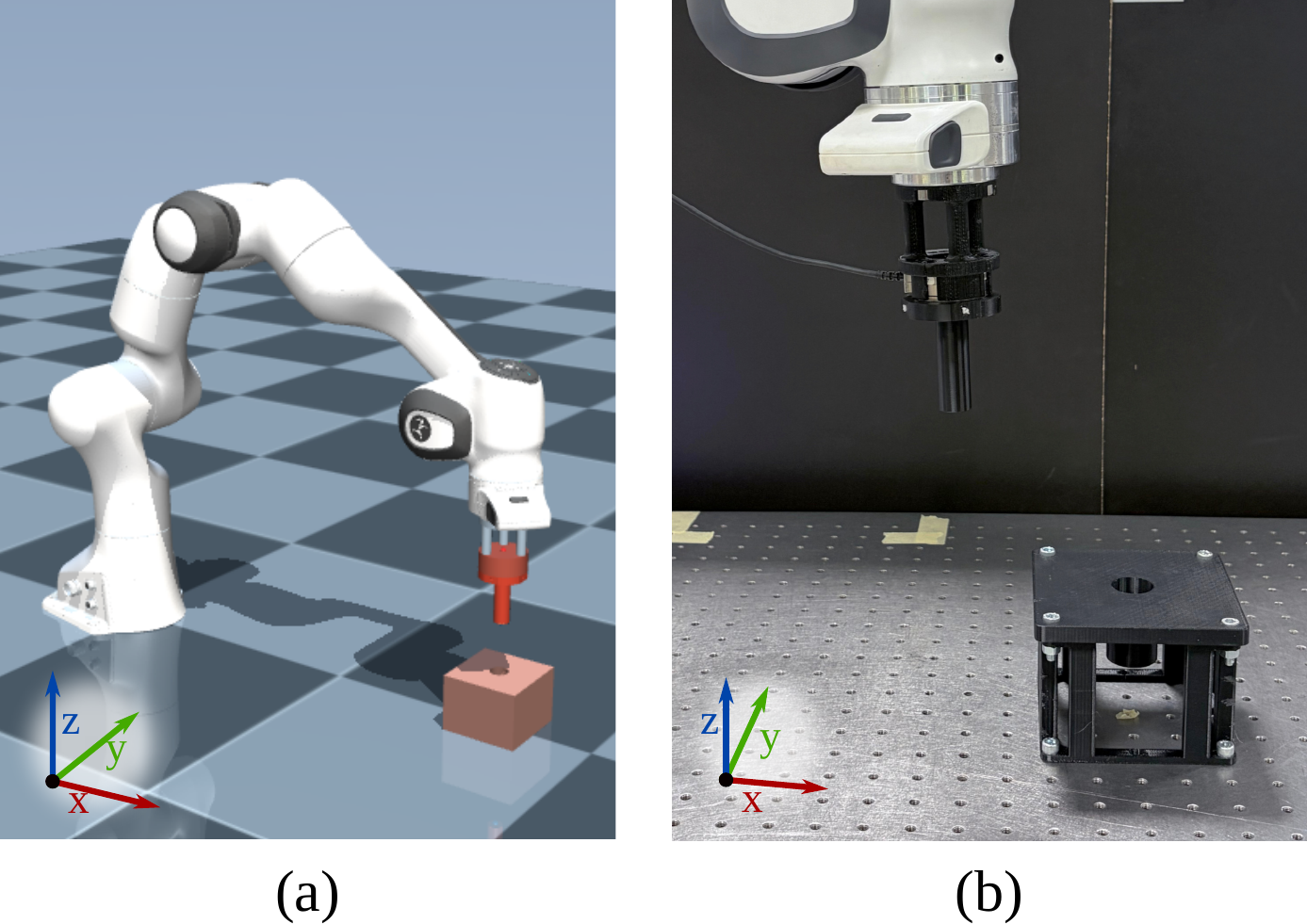}
    \caption{Simulation and real-robot setups for the peg-in-hole task. (a) MuJoCo environment with a Franka Emika Panda, cylindrical peg, and fixed hole. (b) Physical setup with the same robot and peg--hole geometry, instrumented with an ATI force--torque sensor.}
    \label{fig:exp-setup}
\end{figure}

The task is performed by a Franka Emika Panda robot equipped with a force--torque sensor and a cylindrical peg, as shown in \autoref{fig:exp-setup}. The controlled Cartesian frame is attached to the peg tip ($\mathbf{p}_{c,t} \equiv \mathbf{p}_{\mathrm{peg},t}$). The peg and hole diameters are 20~mm and 23.6~mm, respectively. An episode terminates with success when the peg remains inserted for 2 second, or with failure when the timeout is reached.

The simulation environment is implemented in MuJoCo. At the beginning of each training episode, the true insertion goal $\mathbf{p}_g$ is randomized independently along the x-y axes. The policy does not observe $\mathbf{p}_g$ directly, instead, it receives an estimated goal position $\hat{\mathbf{p}}_{g,t}$ with an episode-independent estimation error. Additional noise is applied to the goal-relative position and measured force--torque observations. The randomization ranges are summarized in \autoref{tab:table_domain_randomization}.

\begin{table}[htbp]
\centering
\caption{Task randomization and observation-noise settings.}
\label{tab:table_domain_randomization}
\begin{tabular}{ll}
\toprule
Parameter & Distribution or range \\
\midrule
Initial hole offset $[\Delta x,\Delta y]$ &
$[-50,50]~\mathrm{mm}$ per axis \\

Goal-pos. estimation error $[e_x,e_y]$ &
$\mathcal{N}\!\left(\mathbf{0},(1.5~\mathrm{mm})^2\mathbf{I}\right)$ \\

Goal-distance noise $[n_x,n_y,n_z]$ &
$\mathcal{N}\!\left(
\mathbf{0},
\mathrm{diag}(0.5,0.5,0.3)^2
\right)\, \mathrm{mm}$ \\

Force observation noise &
$\mathcal{N}\!\left(\mathbf{0},(0.2~\mathrm{N})^2\mathbf{I}\right)$ \\

Torque observation noise &
$\mathcal{N}\!\left(\mathbf{0},(0.02~\mathrm{N\,m})^2\mathbf{I}\right)$ \\
\bottomrule
\end{tabular}
\end{table}

\subsubsection{Training Configuration}

The physics simulation and the Cartesian impedance controller run at 1000~Hz. PA-RL updates its attractor-field parameters at every 2~second, while the field is re-evaluated at 100~Hz to produce a bounded Cartesian reference; because this reference is a state-dependent feedback field, the trajectory keeps adapting to the current peg-tip state even when the parameters are held fixed. The TCP-Vel, TCP-Pose, and VICES baselines instead output local motion or impedance parameter and therefore act at the control rate of 62.5~Hz. These rates follow from the temporal semantics of each action interface, not from any computational limitation.

All policies were trained with Soft Actor–Critic using two-hidden-layer $256$ ReLU networks, a batch size of $256$, a target smoothing coefficient $\tau = 0.005$, and one gradient step per environment step. Since the four methods use fundamentally different action representations, a single shared hyperparameter set would favour one method and understate the others; we therefore keep the task, reward, and success criterion identical and tune only the learning-side hyperparameters per method, evaluating PA-RL and baselines at its best-performing configuration so the comparison reflects each representation's maximum attainable performance. PA-RL used a learning rate of $5\times10^{-4}$, $\gamma = 0.6$, a replay buffer of $6\times10^{4}$ policy step, and $888$ warm-up steps; the baselines used $3\times10^{-4}$, $\gamma = 0.995$, a buffer of $2\times10^{5}$, and $3\times10^{4}$ warm-up steps.

\subsubsection{Action Space}
\label{sec:Action Space}

For PA-RL, the APF action representation contains one linear
attractor and two Gaussian attractors. The policy outputs their goal-relative offsets, weights, and Gaussian sigma, resulting in a 14-dimensional action. Attractor offsets are expressed in normalized coordinates and mapped to the goal relative operational space which is defined from the range of initial peg-tip positions and the target insertion region

The offsets range are based on the operation space while weights and sigmas range preserve the linear attractor as the primary component while allowing the Gaussian attractors to locally modify the field. The broadest attractor region covers approximately \(0.8\) of this envelope, while the other two regions provide complementary spatial resolutions within the same workspace rather than the full robot workspace. The linear attractor weight is bounded by \(w_1\in[0.20,0.80]\), while the Gaussian attractor weights satisfy \(w_2,w_3\in[0,0.12]\). Their Gaussian sigmas are \(\sigma_2\in[20,80]\) and \(\sigma_3\in[8,30]\). These bounds retain the linear attractor as the principal goal-directed component while allowing the Gaussian attractors to reshape the field online from geometric and force--torque observations.

TCP-Vel directly outputs a three-dimensional peg-tip velocity, bounded
by
$\|\mathbf{v}_t\|_2 \leq 0.03~\mathrm{m/s}$.
TCP-Pose outputs a three-dimensional Cartesian position increment, with
each component constrained to $[-2,2]$~mm. The VICES-style
variable-impedance baseline extends the position increment with
translational stiffness and damping-ratio commands, resulting in a
nine-dimensional action. Its translational stiffness is bounded by
$[200,200,250]^\top$ and $[800,800,900]^\top$~N/m, while the damping
ratio is bounded by $[0.70,0.70,0.80]^\top$ and
$[1.60,1.60,1.85]^\top$. The rotational impedance remains fixed.

\subsubsection{Observation Space}
\label{sec:Observation Space}
All policies receive the goal-relative position
$\hat{\mathbf{p}}_{g,t}-\mathbf{p}_{\mathrm{peg},t}$ and the measured
force--torque signal. PA-RL additionally observes the current
attractor offsets, producing an 18-dimensional observation. TCP-Vel and
TCP-Pose additionally observe the peg-tip velocity and previous action,
producing 15-dimensional observations. VICES further observes its
current normalized translational stiffness and damping ratio, producing
a 27-dimensional observation. The observation dimensions are not
artificially equalized because each interface requires access to its
own execution state.

\subsubsection{Training and Evaluation}
\label{sec:training_evaluation_protocol}

All policies are trained using SAC algorithm with three random
seeds. They share the same MuJoCo model, task reward, insertion-success
criterion, and force--torque observation model. The configured training budgets are $5\times10^{6}$ environment steps for all the methods.

Let
\begin{equation}
d_t =
\left\|
\mathbf{p}_{\mathrm{peg},t}
-
\mathbf{p}_{g}
\right\|_2
\end{equation}
denote the distance from the peg tip to the true insertion goal. The
shared reward is
\begin{equation}
r_t =
10\left(1-\tanh(45d_t)\right)
+
60\,\mathbb{I}[s_t=1],
\label{eq:reward_function}
\end{equation}
where $s_t$ is the insertion-success indicator. The true insertion goal is used only for reward computation and evaluation.

Policies are evaluated periodically on a common fixed $3\times3$ grid, with horizontal goal offsets
$x,y\in\{-50,0,50\}$~mm. Each position is evaluated three times, resulting in 27 episodes per checkpoint. The true goal position remains fixed across the three trials at each grid point, while the goal-position estimation error is resampled at every episode and observation noise is resampled independently during execution. The configured random seeds are shared across methods.

\subsection{Experiment Results}

We report various metrics to evaluate the learning efficiency, task performance and motion quality, as described below.

\begin{figure}[htbp]
    \centering
    \includegraphics[width=\linewidth]{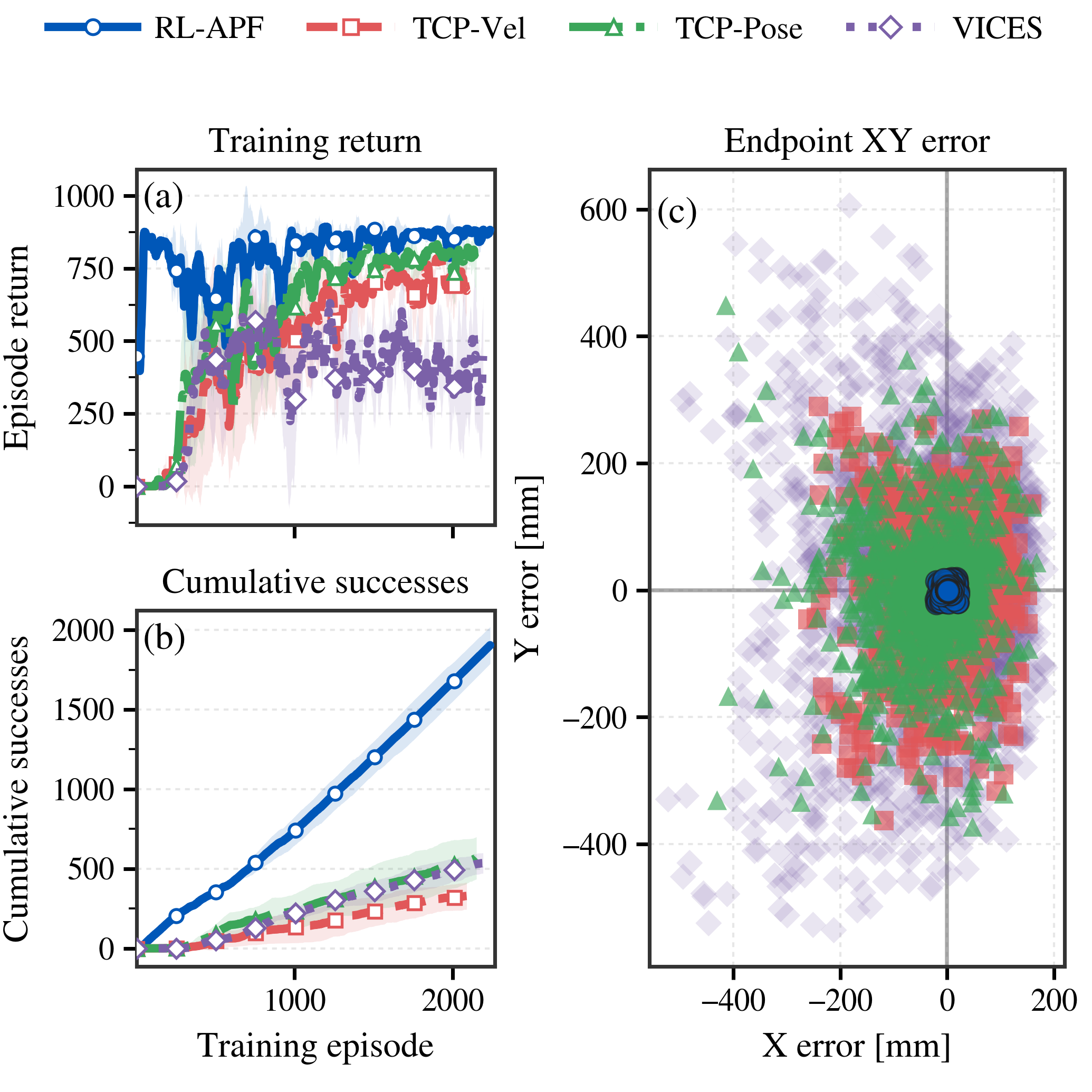}
    \caption{Learning efficiency over three seeds. Solid lines and shaded regions denote the across-seed mean and one standard deviation. The endpoint distribution expressed as the final peg-tip XY error relative to the true insertion goal.}
    \label{fig:learning_efficiency}
\end{figure}

\textbf{\textit{Learning efficiency}} is assessed using episode return, cumulative successful episodes, and the episode endpoint distribution as presented in \autoref{fig:learning_efficiency}. PA-RL exhibits an rapid increase in episode return and maintains a higher return throughout most of the common training range. Its cumulative-success curve also grows substantially faster, indicating that successful insertion behavior is acquired earlier and reproduced more frequently during training. In contrast, the baseline methods improve more gradually and show greater variation in their return curves.

The endpoint distribution provides complementary spatial evidence. PA-RL produces a compact endpoint cluster around the true hole center, whereas the baseline distributions are more dispersed endpoints. These distant samples are consistent with incomplete approaches, residual misalignment, or termination before insertion. These results indicate earlier episode-level acquisition of task-relevant behavior.

\textbf{\textit{Task performance}} is assessed using evaluation success rate and the mean final peg-tip-to-goal distance with the $N{=}27$ fixed-grid evaluation episodes:

\begin{equation}
d_{\mathrm{final}}
=
\frac{1}{N}
\sum_{i=1}^{N}
\left\|
\mathbf{p}^{(i)}_{\mathrm{peg},T_i}
-
\mathbf{p}^{(i)}_{g}
\right\|_2 ,
\label{eq:final_distance}
\end{equation}
where $s_i$ is the success indicator, $T_i$ is the final sample of
episode $i$, and $\mathbf{p}^{(i)}_{g}$ is the corresponding true
insertion goal. 

\begin{figure}[htbp]
    \centering
    \includegraphics[width=1\linewidth]
    {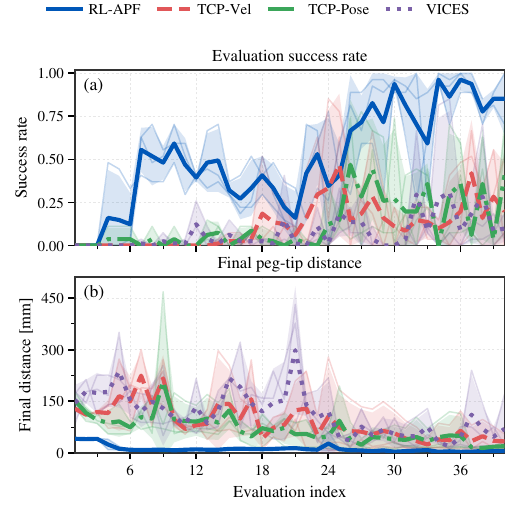}
    \caption{Fixed-grid evaluation task performance. Each evaluation process contains 27 episodes with nine hole positions on a \(3\times3\) grid and three trials per position under resampled goal-position estimation error and observation noise. Solid lines and shaded regions denote the across-seed mean and one standard deviation.}
    \label{fig:task_performance}
\end{figure}

As shown in \autoref{fig:task_performance}, PA-RL is the first and only method to reach a 100\% evaluation success rate. The final peg-tip-to-goal distance to approximately $2$~mm. In contrast, the baseline methods occasionally reach high success rates at individual checkpoints but do not sustain this performance consistently. Their success curves exhibit stronger fluctuations and larger cross-seed variation, accompanied by repeated spikes in the final-distance curves. This pattern indicates that their failures are not merely caused by narrowly missing the success threshold. Instead, a considerable proportion of episodes terminate with incomplete insertion or substantial residual misalignment.

\textbf{\textit{Motion quality}} is assessed using peak measured interaction force, RMS inter-step joint-torque variation, path length, and RMS Cartesian acceleration. Motion quality metrics are averaged over all 27 evaluation episodes, rather than successful episodes only.

For evaluation episode $i$, the Cartesian path length is
\begin{equation}
L_{\mathrm{path}}^{(i)}
=
\sum_{t=2}^{T_i}
\left\|
\mathbf{p}^{(i)}_{\mathrm{peg},t}
-
\mathbf{p}^{(i)}_{\mathrm{peg},t-1}
\right\|_2 .
\label{eq:path_length}
\end{equation}

The peak measured interaction force is defined as
\begin{equation}
f_{\mathrm{peak}}^{(i)}
=
\max_{1\leq t\leq T_i}
\left\|
\mathbf{f}^{(i)}_t
\right\|_2  ,
\label{eq:peak_force}
\end{equation}
where $\mathbf{f}^{(i)}_t\in\mathbb{R}^3$ is the force component of
the measured force--torque signal.

RMS inter-step joint-torque variation is computed as
\begin{equation}
D_{\tau}^{(i)}
=
\frac{1}{T_i-1}
\sum_{t=2}^{T_i}
\sqrt{
\frac{1}{J}
\sum_{j=1}^{J}
\left(
\tau^{(i)}_{t,j}
-
\tau^{(i)}_{t-1,j}
\right)^2
},
\label{eq:torque_variation}
\end{equation}
where $J=7$ is the number of robot joints.

RMS Cartesian acceleration (velocity roughness) is computed from consecutive peg-tip
velocities as
\begin{equation}
A_{\mathrm{RMS}}^{(i)}
=
\sqrt{
\frac{1}{T_i-1}
\sum_{t=2}^{T_i}
\left\|
\frac{
\dot{\mathbf{p}}_{c,t}^{(i)}
-
\dot{\mathbf{p}}_{c,t-1}^{(i)}
}{
\Delta t
}
\right\|_2^2
}.
\label{eq:rms_acceleration}
\end{equation}

\begin{figure}[htbp]
    \centering
    \includegraphics[width=1\linewidth]{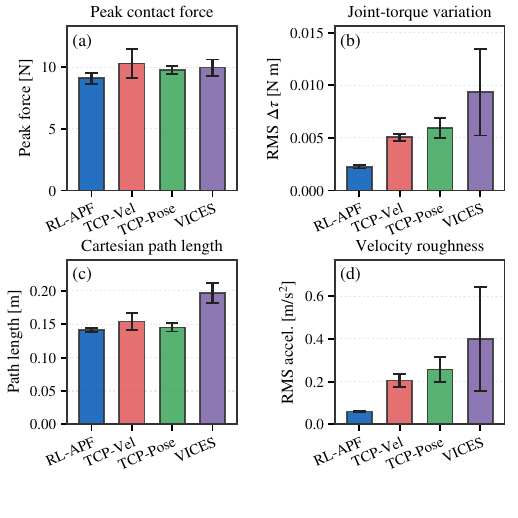}
    \caption{Motion quality at the selected evaluation experiment. For each method and seed, the checkpoint with the highest evaluation success rate is selected. Metrics are computed over all 27 evaluation episodes. Bars and error bars denote the across-seed mean and one standard deviation; markers denote individual seeds.}
    \label{fig:motion_quality}
\end{figure}

Each motion-quality result reported in Fig.~\ref{fig:motion_quality} is the mean of the corresponding per-episode metric over all $N=27$ fixed-grid evaluation episodes:
\begin{equation}
\bar{m}
=
\frac{1}{N}
\sum_{i=1}^{N}m^{(i)},
\qquad
m^{(i)}
\in
\left\{
L_{\mathrm{path}}^{(i)},
f_{\mathrm{peak}}^{(i)},
D_{\tau}^{(i)},
A_{\mathrm{RMS}}^{(i)}
\right\}.
\label{eq:metric_average}
\end{equation}

The results in \autoref{fig:motion_quality} reflect distinct execution behaviors induced by the policy action representation. PA-RL maintained a comparable peak measured interaction force of \(9.10\pm0.45~\mathrm{N}\), while producing lower RMS inter-step joint-torque variation (\(0.00226\pm0.00016~\mathrm{N\,m}\)) and Cartesian acceleration (\(0.0599\pm0.0016~\mathrm{m/s^2}\)) than the baselines. Its trajectories exhibited continuous goal-directed motion and gradual contact correction, whereas the direct motion-command interfaces more frequently revised their commands near the hole.

Although TCP-Pose achieved a similar path length, its execution contained larger local variations, showing that path length alone does not capture motion regularity. Unsuccessful baseline rollouts typically terminated near the hole after repeated correction or sustained contact without completing insertion. As the common reward contains no explicit motion quality penalties, these behaviors are consistent with the different structures introduced by the policy interfaces rather than method-specific reward shaping.

Overall, PA-RL achieved faster learning, stronger task performance under added estimation error, and smoother robot motion.

\subsection{Real-world Experiment}

\begin{figure}[htbp]
    \centering
    \includegraphics[width=\linewidth]{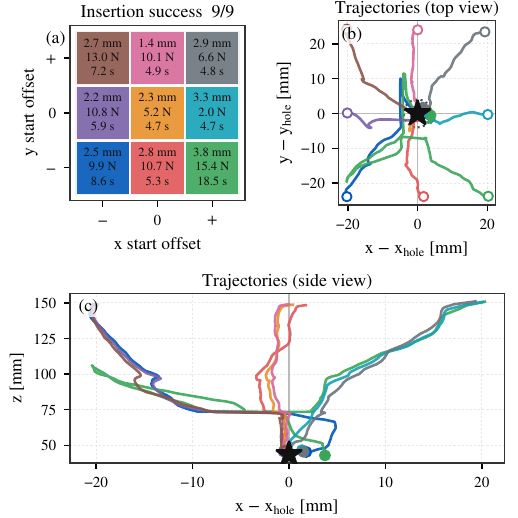}
    \caption{Real-robot deployment over a $3{\times}3$ grid. (a) Per-trial task performance with final peg-tip-to-goal distance, peak measured contact force, and completion time. (b) Top-view peg-tip trajectories relative to the true insertion goal and (c) Side-view trajectories. Colors indicate the different initial relative peg--hole positions.}
    \label{fig:real_experiment}
\end{figure}

The simulation-trained PA-RL policy was deployed on the physical Franka robot (\autoref{fig:exp-setup}(b)) without policy fine-tuning. Across the $3{\times}3$ start-offset grid, it completed 9/9 peg-in-hole insertions and the results are shown in \autoref{fig:real_experiment}. The mean final peg-tip-to-goal distance was 2.66 mm, with a range of 1.4–3.8 mm. The mean peak measured interaction force was 9.3 N (2.0–15.4 N), and the mean completion time was 7.18 s (4.7–18.5 s). 

The top-view trajectories approach the insertion goal from different lateral directions and converge within the hole region rather than following a single prescribed path. The side-view trajectories in~\autoref{fig:real_experiment}(c) show lateral correction during approach followed by descent along the insertion axis, showing the trajectory variations that are caused by the different initial offsets and contact sequences. The real-robot experiments demonstrate deployment feasibility.

Qualitatively, the trajectories and the supplementary video show contact-dependent adaptation. The policy does not simply execute a fixed downward insertion motion; instead, it performs lateral corrections around the hole and, in some trials, briefly moves upward and reattempts insertion multiple times.
This suggests that the learned APF representation can exploit force--torque feedback and peg--hole alignment cues to guide exploratory sliding near the hole before successful insertion. Thus, the real-robot experiment demonstrates both deployment feasibility and adaptive interaction behavior.

\section{Conclusion}

This work introduced PA--RL, a framework that allows an RL policy to adapt an APF as an interpretable and spatially structured motion representation. The method uses model-free RL for task-level adaptation, while a bounded APF-based reference generator and a fixed Cartesian impedance controller handle continuous low-level motion generation and tracking. This separates learned field shaping from direct motion-command execution, reducing the burden on the policy.

The results show that PA--RL improves learning efficiency, task performance, and motion quality compared with direct motion-command and variable-impedance baselines. In simulation, PA--RL achieved faster learning, stronger task performance under goal-position estimation error, and smoother robot motion. The simulation-trained policy also completed all real-robot insertions without policy fine-tuning, suggesting robustness to the sim-to-real gap and supporting the feasibility of the learned potential-field interface.

At the same time, the present work should be viewed as an early instantiation rather than a general solution. While APFs are broadly applicable, the current parameterization includes task-relevant components, such as goal-relative attractors, that are well suited to peg-in-hole insertion. Moreover, the current formulation does not guarantee global convergence to the task goal. Future work will investigate richer APF parameterizations, broader manipulation scenarios, systematic ablations, stronger convergence analysis, improved sensory representations, and larger-scale real-robot evaluations.

\addtolength{\textheight}{-5cm}   






\bibliographystyle{IEEEtran}
\bibliography{references}

\end{document}